\documentclass[lettersize,journal]{IEEEtran}  % Comment this line out if you need a4paper

\IEEEoverridecommandlockouts                              % This command is only needed if 
\usepackage{graphics} % for pdf, bitmapped graphics files
\usepackage{epsfig} % for postscript graphics files
\usepackage{times} % assumes new font selection scheme installed
\usepackage{amsmath} % assumes amsmath package installed
\usepackage{amssymb}  % assumes amsmath package installed
\usepackage{gensymb}
\usepackage{multirow}
\usepackage{hhline}
\usepackage{booktabs}
\usepackage{float}
\usepackage{array}
\usepackage{tabularray}
\usepackage{makecell}
\usepackage[font=small,labelfont=bf]{caption}
\newcommand{\myparagraph}[1]{\vspace{0.05in}\noindent\textbf{#1}}
\usepackage{hyperref}
\hypersetup{
    colorlinks=true,
    linkcolor=blue,
    filecolor=magenta,      
    urlcolor=cyan,
    pdftitle={Overleaf Example},
    pdfpagemode=FullScreen,
    }

\usepackage{caption}% http://ctan.org/pkg/caption
\usepackage{cite}
\title{\bf
\LARGE TEXterity - Tactile Extrinsic deXterity\\
\large Simultaneous Tactile Estimation and Control for Extrinsic Dexterity
}

\author{
  \IEEEauthorblockN{$^*$Sangwoon Kim$^{1}$, $^*$Antonia Bronars$^{1}$, Parag Patre$^{2}$ and Alberto Rodriguez$^{1}$}\\
  \IEEEauthorblockA{
     $^*$Equal Contribution, $^{1}$MIT, $^{2}$Magna International Inc.\\
     {\tt\small <sangwoon,bronars,albertor>@mit.edu, parag.patre@magna.com}
     }
}

\begin{document}

\twocolumn[{%
\renewcommand\twocolumn[1][]{#1}%
\maketitle
\begin{center}
    \centering
    \captionsetup{type=figure}
    \vspace{-5mm}
    \includegraphics[width=.9\textwidth]{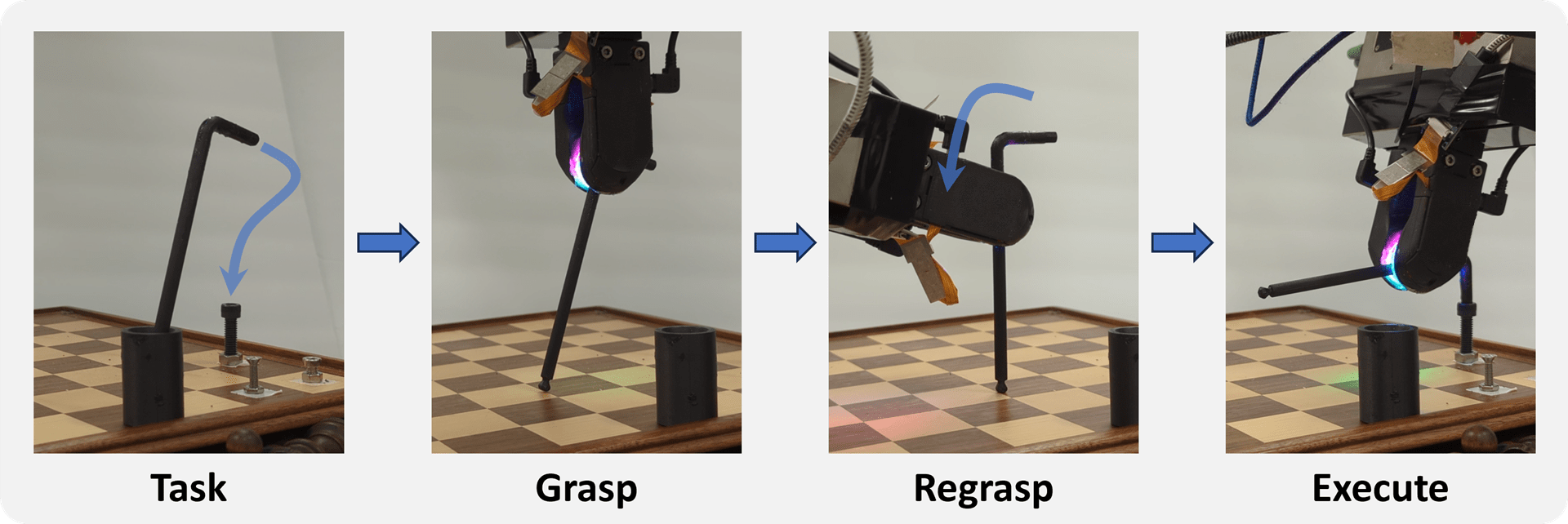}
    \captionof{figure}{An example task that requires tactile extrinsic dexterity. A proper grasp is essential when using an Allen key to apply sufficient torque while fastening a hex bolt. The proposed method utilizes tactile sensing on the robot's finger to localize and track the grasped object's pose and also regrasp the object in hand by pushing it against the floor - effectively leveraging extrinsic dexterity.}
    \label{fig:first}
\end{center}%
}]

\thispagestyle{empty}
\pagestyle{empty}

\begin{abstract}

We introduce a novel approach that combines tactile estimation and control for in-hand object manipulation. By integrating measurements from robot kinematics and an image-based tactile sensor, our framework estimates and tracks object pose while simultaneously generating motion plans to control the pose of a grasped object. This approach consists of a discrete pose estimator that tracks the most likely sequence of object poses in a coarsely discretized grid, and a continuous pose estimator-controller to refine the pose estimate and accurately manipulate the pose of the grasped object. Our method is tested on diverse objects and configurations, achieving desired manipulation objectives and outperforming single-shot methods in estimation accuracy. The proposed approach holds potential for tasks requiring precise manipulation and limited intrinsic in-hand dexterity under visual occlusion, laying the foundation for closed-loop behavior in applications such as regrasping, insertion, and tool use. Please see \href{https://sites.google.com/view/texterity}{this url} for videos of real-world demonstrations.

\end{abstract}

%%%%%%%%%%%%%%%%%%%%%%%%%%%%%%%%%%%%%%%%%%%%%%%%%%%%%%%%%%%%%%%%%%%%%%%%%%%%%%%%
\section{INTRODUCTION}
The ability to manipulate objects within the hand is a long-standing objective in robotics for its potential to increase the workspace, speed, and capability of robotic systems. For example, the ability to change the grasp on an object can improve grasp stability and functionality, or prevent collisions and kinematic singularities. In-hand manipulation is challenging from the perspectives of state estimation, planning, and control: first, once the object is enveloped by the grasp, it becomes difficult to perceive with external vision systems; second, the hybrid dynamics of contact-rich tasks are difficult to predict \cite{bauza2018data} and optimize over \cite{mordatch2012contact}.

Existing work on in-hand manipulation emphasizes the problem of sequencing contact modes, and can be broken down into two prevailing methodologies. One line of work relies on simple object geometries and exact models of contact dynamics to plan using traditional optimization-based approaches \cite{sundaralingam2018geometric, mordatch2012contact, hou2018fast, shi2017dynamic, sundaralingam2019relaxed}, while the other leverages model-free reinforcement learning to learn policies directly that only consider or exploit contact modes implicitly \cite{chen2022system, rajeswaran2017learning, chen2023visual, handa2023dextreme, andrychowicz2020learning, huang2021generalization}. Much less consideration has been given to the challenge of precisely controlling such behaviors, despite the fact that prominent tasks like connector insertion or screwing in a small bolt require high precision. 

Tactile feedback is a promising modality to enable precise control of in-hand manipulation. Image-based tactile sensing \cite{yuan2017gelsight, lambeta2020digit, taylor2022gelslim} has gained traction in recent years for its ability to provide high-resolution information directly at the contact interface. Image-based tactile sensors have been used for pose estimation \cite{bauza2022tac2pose}, object retrieval \cite{pai2023tactofind}, and texture recognition \cite{luo2018vitac}. They have also been used to estimate the location of contacts with the environment \cite{kim2022active, ma2021extrinsic, higuera2023neural}, to supervise insertion \cite{dong2021tactile}, and to guide the manipulation of objects like boxes \cite{hogan2020tactile}, tools \cite{shirai2023tactile}, cable \cite{she2021cable}, and cloth \cite{sunil2023visuotactile}.

We study the problem of precisely controlling in-hand sliding regrasps by pushing against an external surface, i.e. extrinsic dexterity \cite{dafle2014extrinsic}, supervised only by robot proprioception and tactile sensing. Our framework is compatible with arbitrary, but known, object geometries and succeeds even when the contact parameters are known only approximately.

This work builds upon previous research efforts. First, \textit{Tac2Pose} \cite{bauza2022tac2pose} estimates the relative gripper/object pose using tactile sensing, but lacks control capabilities. Second, \textit{Simultaneous Tactile Estimation and Control of Extrinsic Contact} \cite{kim2023simultaneous} estimates and controls extrinsic contact states between the object and its environment, but has no understanding of the object's pose and therefore has limited ability to reason over global re-configuration. Our approach combines the strengths of these two frameworks into a single system. As a result, our method estimates the object's pose and its associated contact configurations and simultaneously controls them. By merging these methodologies, we aim to provide a holistic solution for precisely controlling general planar in-hand manipulation.

This paper is an extension of our work on \textit{tactile extrinsic dexterity} \cite{bronars2024texterity} in these ways:
\begin{itemize}
    \item In Section \ref{subsection:varigoal}, we evaluate our method against five ablations for four distinct types of goal configurations. These new results illuminate key features of our approach. In particular, we evaluate the effectiveness of leveraging prior knowledge of the external environment to collapse ambiguity in individual tactile images. In addition, we compare our results against those derived from idealized simulations and using privileged information, to showcase the capability of our approach in bridging the sim-to-real gap.
    \item In Section \ref{subsection:household}, we provide qualitative results for three household objects in realistic scenarios. These results motivate the work concretely, and demonstrate that our method generalizes to real objects, which have a variety of material, inertial, and frictional properties.
    \item Finally, we provide a more complete review of prior work in Section \ref{section:relatedwork}, and more thorough explanation of our method in Sections \ref{subsection:discrete} and \ref{subsection:continuous}.
\end{itemize}
%, without introducing additional sensors.
%(rendered contact images, and rendered contact images + idealized physics).
%We find that the estimation accuracy does not differ significantly when using real vs. rendered contact images. This comparison demonstrates that our approach, which relies on perception models trained solely in simulation and analytical physics, successfully overcomes the sim-to-real gap.

\begin{figure*}[h]
	\centering
	\includegraphics[width=1.0\linewidth]{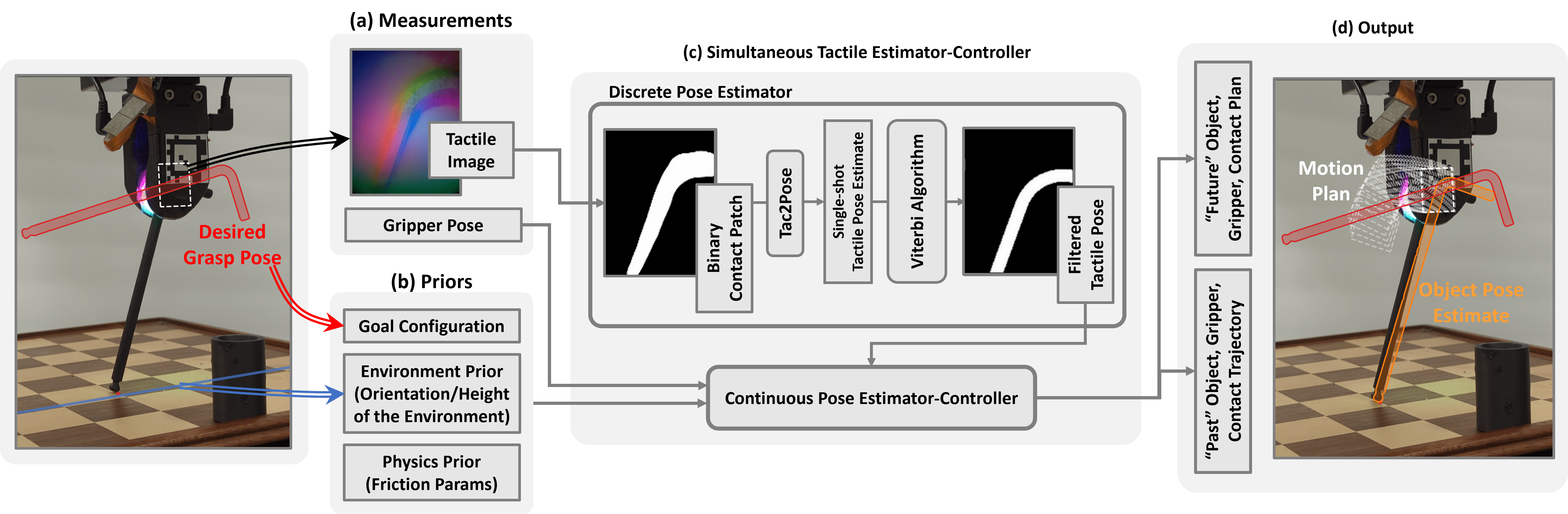}
	\caption{Overview of the Simultaneous Tactile Estimation and Control Framework.}
	\label{fig:overview}
\end{figure*}

\section{RELATED WORK} \label{section:relatedwork}
\myparagraph{Tactile Estimation and Control.} Image-based tactile sensors are particularly useful for high-accuracy pose estimation, because they provide high-resolution information about the object geometry throughout manipulation. They have been successfully used to track object drift from a known initial pose \cite{sodhi2021learning, sodhi2022patchgraph}, build a tactile map and localize the object within it \cite{zhao2023fingerslam, suresh2021tactile, bauza2019tactile}, and estimate the pose of small parts from a single tactile image \cite{li2014localization}. Because touch provides only local information about the object geometry, most tactile images are inherently ambiguous \cite{bauza2022tac2pose}. Some work has combined touch with vision \cite{bauza2023simple, dikhale2022visuotactile, izatt2017tracking, anzai2020deep} to resolve such ambiguity. Our approach is most similar to a line of work that estimates distributions over possible object pose from a single tactile image \cite{bauza2022tac2pose, suresh2023midastouch, kelestemur2022tactile}, then fuses information over streams of tactile images using particle \cite{suresh2023midastouch} or histogram \cite{kelestemur2022tactile} filters. \cite{suresh2023midastouch} tackles the estimation, but not control, problem, assuming that the object is rigidly fixed in place while a human operator slides a tactile sensor along the object surface. Similarly, \cite{kelestemur2022tactile} also assumes the object is fixed in place, while the robot plans and executes a series of grasp and release maneuvers to localize the object. Our work, on the other hand, tackles the more challenging problem of estimating and controlling the pose of an object sliding within the grasp while not rigidly attached to a fixture. The mechanics of sliding on a deformable sensor surface are difficult to predict, which places more stringent requirements on the quality of the observation model and controller.

\myparagraph{In-Hand Manipulation.} In-hand manipulation is most commonly achieved with dexterous hands or by leveraging the surrounding environment (extrinsic dexterity \cite{dafle2014extrinsic}). One line of prior work formulates the problem as an optimization over exact models of the hand/object dynamics \cite{sundaralingam2018geometric, mordatch2012contact, hou2018fast, shi2017dynamic, sundaralingam2019relaxed, hou2020manipulation}, but only for simple objects and generally only in simulation \cite{sundaralingam2018geometric, mordatch2012contact}, or by relying on accurate knowledge of physical parameters to execute plans precisely in open loop \cite{hou2018fast}.
Another line of prior work focuses on modeling the mechanics of contact itself in a way that is useful for planning and control, either analytically \cite{shi2020hand, chavan2018hand, chavan2015prehensile} or with neural networks \cite{nagabandi2020deep, kumar2016optimal}.

Some work has avoided the challenges of modeling contact altogether, instead relying on model-free reinforcement learning with vision to directly learn a policy for arbitrary geometries. Some policies have been tested on simulated vision data only \cite{chen2022system, rajeswaran2017learning}, while others operate on real images \cite{chen2023visual, handa2023dextreme, andrychowicz2020learning, huang2021generalization}. They, however, suffer from a lack of precision. As an example, \cite{chen2023visual} reports 45$\%$ success on held out objects, and 81$\%$ success on training objects, where success is defined as a reorientation attempt with less than 0.4 rad (22.9\degree) of error, underscoring the challenge of precise reorientation for arbitrary objects.

There have also been a number of works leveraging tactile sensing for in-hand manipulation. \cite{she2021cable}, \cite{sunil2023visuotactile}, and \cite{tian2019manipulation} use image-based tactile sensors to supervise sliding on cables, cloth, and marbles, respectively. \cite{shirai2023tactile} detects and corrects for undesired slip during tool manipulation, while \cite{lepert2023hand} learns a policy that trades off between tactile exploration and execution to succeed at insertion tasks. Some works rely on proprioception \cite{pitz2023dextrous} or pressure sensors \cite{van2015learning} to coarsely reorient objects within the hand. State estimation from such sensors is challenging and imprecise, leading to policies that accrue large errors. Another line of work uses tactile sensing to reorient objects within the hand continuously \cite{qi2023general, khandate2023sampling, yin2023rotating, sievers2022learning, yuan2023robot}, without considering the challenge of stopping at goal poses precisely.

We consider the complementary problem of planning and controlling over a known contact mode (in-hand sliding by pushing the object against an external surface), where the object geometry is arbitrary but known. We leverage a simple model of the mechanics of sliding and supervise the behavior with high-resolution tactile sensing, in order to achieve precise in-hand manipulation. By emphasizing the simultaneous estimation and control for a realistic in-hand manipulation scenario, this work addresses a gap in the existing literature and paves the way for executing precise dexterous manipulation on real systems.

\myparagraph{Extrinsic Contact Estimation and Control.} Extrinsic contacts, or contacts between a grasped object and the surrounding environment, are fundamental to a range of contact-rich tasks including insertion, tool use, and in-hand manipulation via extrinsic dexterity. A variety of work has explored the ability to estimate \cite{ma2021extrinsic, higuera2023neural, van2023integrated} and control \cite{doshi2022manipulation, kim2022active, kim2023simultaneous,higuera2023perceiving} such contacts using intrinsic (on the robot) sensing. 

\cite{doshi2022manipulation} manipulates unknown objects by estimating and controlling extrinsic contacts with force-torque feedback. \cite{ma2021extrinsic} uses image-based tactile feedback with a small exploratory motion to localize an extrinsic point contact that is fixed on the environment. \cite{kim2022active, kim2023simultaneous} estimates and controls extrinsic contacts represented as points, lines, and patches with feedback from image-based tactile sensors.

Another line of prior work instead represents and estimates extrinsic contacts using neural implicit functions with tactile \cite{higuera2023neural, higuera2023perceiving} or visuo-tactile \cite{van2023integrated} sensing. Finally, \cite{kim2023im2contact} estimates extrinsic contacts from a scene-level RGB-D images of the robot workspace. These methods are complementary to our approach, which explicitly represents the extrinsic contacts using a kinematic model, rather than using implicit neural representations of the extrinsic contacts.

\section{METHOD}

\begin{figure*}[h]
	\centering
	\includegraphics[width=1.0\linewidth]{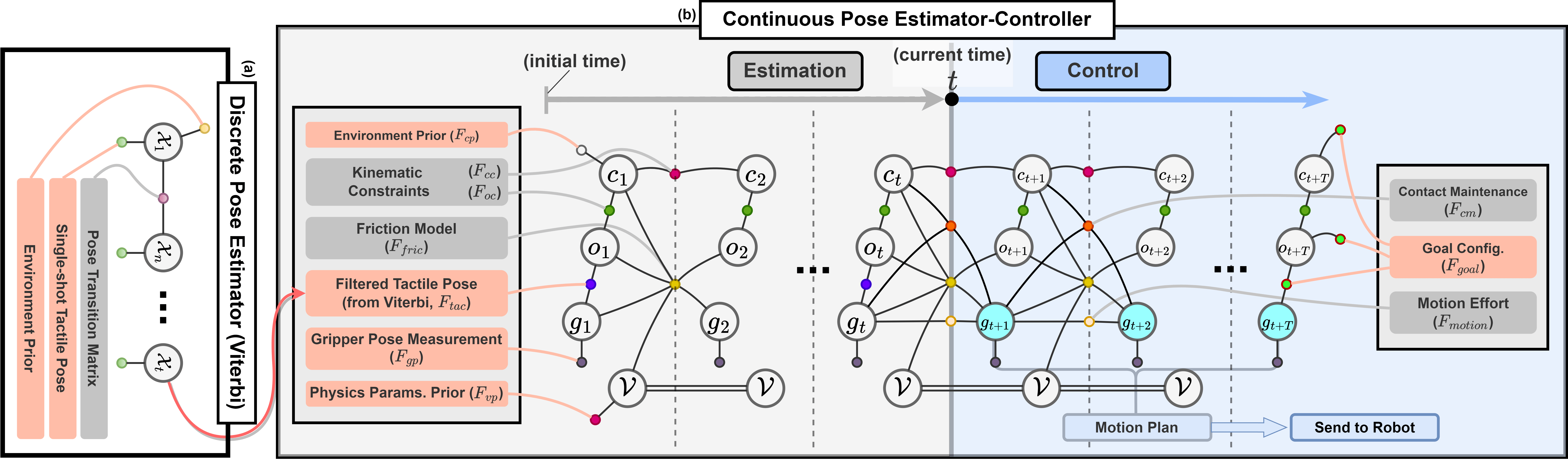}
	\caption{Graph Architecture of the Simultaneous Tactile Estimator-Controller.}
	\label{fig:graph}
\end{figure*}

\subsection{Problem Formulation}
\label{formulation}

We address the task of manipulating objects in-hand from unknown initial grasps to achieve desired configurations by pushing against the environment. The target configurations encompass a range of potentially simultaneous manipulation objectives:

\begin{itemize}
    \item Changing the grasp pose (i.e., relative rotation/translation between the gripper and the object)
    \item Changing the orientation of the object in the world frame (i.e., pivoting against the environment)
    \item Changing the location of the extrinsic contact point (i.e., sliding against the environment)
\end{itemize}
A wide variety of regrasping tasks can be specified via a combination of the above objectives.

We make several assumptions to model this problem:

\begin{itemize}
    \item Grasped objects are rigid with known 3D models.
    \item The part of the environment that the object interacts with is flat, with a known orientation and height.
    \item Contact between the grasped objects and the environment occurs at a single point.
    \item Grasp reorientation is constrained to the plane of the gripper finger surface.
\end{itemize}

\subsection{Overview} \label{subsection:overview}

Fig. \ref{fig:first} illustrates our approach through an example task: using an Allen key to apply sufficient torque while fastening a hex bolt. Adjusting the grasp through in-hand manipulation is necessary to increase the torque arm and prevent the robot from hitting its motion limit during the screwing.

Fig. \ref{fig:overview} provides an overview of the framework of our approach. The system gathers measurements from both the robot and the sensor (Fig.\ref{fig:overview}a). Robot proprioception provides the gripper's pose, while the GelSlim 3.0 sensor \cite{taylor2022gelslim} provides observation of the contact interface between the gripper finger and the object in the form of an RGB tactile image. The Apriltag attached to the gripper is solely employed for calibration purposes during the quantitative evaluation in Section \ref{subsection:varigoal} and is not utilized as input to the system. The framework also takes as input the desired goal configuration and estimation priors (Fig.\ref{fig:overview}b):

\begin{itemize}
    \item \textbf{Desired Goal Configuration}: A combination of the manipulation objectives discussed in Section \ref{formulation}.
    \item \textbf{Physics Parameter Priors}: The friction parameters at both the intrinsic contact (gripper/object) and the extrinsic contact (object/environment). These priors do not need to be accurate and are manually specified based on physical intuition.
    \item \textbf{Environment Priors}: The orientation and height of the environment in the world frame.
\end{itemize}

Utilizing these inputs, our \textbf{simultaneous tactile estimator-controller} (Fig.\ref{fig:overview}c) calculates pose estimates for the object, along with a motion plan to achieve the manipulation objectives  (Fig.\ref{fig:overview}d). This updated motion plan guides the robot's motion. The framework comprises two main components: \textbf{discrete pose estimator} and \textbf{continuous pose estimator-controller}, which are described in the next subsections.

\subsection{Discrete Pose Estimator} \label{subsection:discrete}

\begin{figure}[h]
	\centering
	\includegraphics[width=1\linewidth]{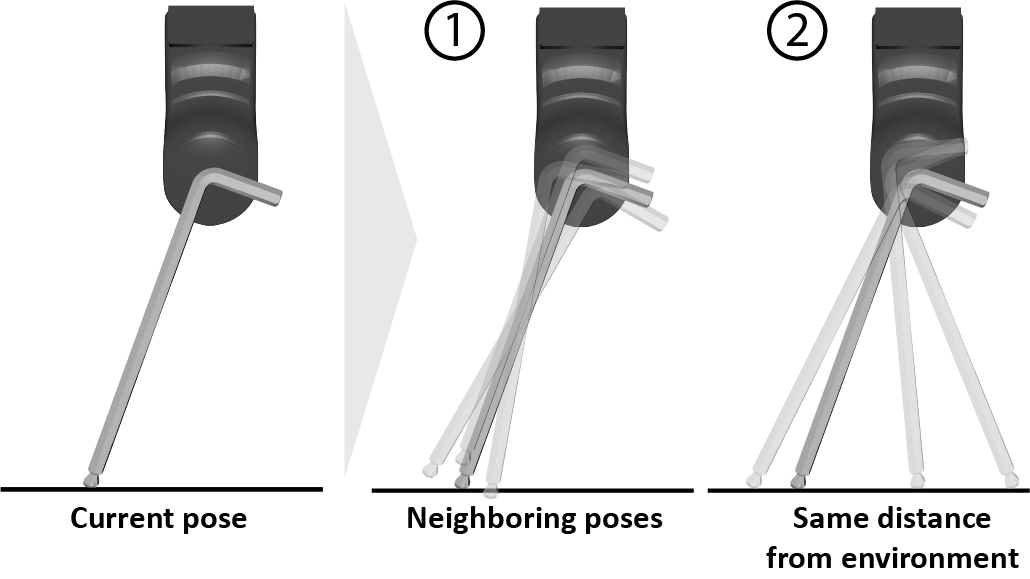}
	\caption{Sample set of allowable transitions on Allen key. The object relative to the gripper finger at the current timestep is shown at left. Possible transitions to new poses at the next timestep are shown at right and center. Transitions favored by the first transition likelihoods (neighboring poses) are shown at center, while those favored by the second transition likelihoods (same distance from the ground) are shown at right.}
	\label{fig:viz_trasitions}
\end{figure}

The discrete pose estimator computes a probability distribution within a discretized grid of relative gripper/object poses. We first describe the process to create pose distributions from single tactile images, and then how to filter through streams of theses distribution estimates.

The individual tactile images are processed as in Tac2Pose \cite{bauza2022tac2pose}. We first reconstruct a binary mask over the region of contact from raw RGB tactile images using a pixel-to-pixel convolutional neural network (CNN) model as described in \cite{bauza2022tac2pose}. Subsequently, the binary mask is channeled into the Tac2Pose estimator \cite{bauza2022tac2pose}, which generates a distribution over possible object poses from a single contact mask.

The Tac2Pose estimator is trained per-object in simulation with rendered contact masks, then transferred directly to the real world. The process for rendering contact masks given an object CAD model is described in detail in \cite{bauza2022tac2pose}. We design a domain randomization procedure tailored for tactile images to ease sim-to-real transfer. These include randomly removing border pixels, tilting the object into and out of the plane of the sensor, randomizing the penetration depth, and randomly removing a fraction of the bottom portion of the sensor (to simulate finger flexing that often occurs during grasping). Once trained, Tac2Pose estimator can run at approximately 50Hz.

We then merge the stream of tactile information with the environment prior via discrete filtering, yielding a filtered probability distribution of the relative object pose. We implement the discrete filter with PGMax \cite{zhou2022pgmax}, running parallel belief propagation for a number of iterations corresponding to the number of variable nodes in the discrete graph. This procedure includes (with some redundant computation) the same belief propagation steps as the Viterbi algorithm \cite{forney1973viterbi}, a standard algorithm for discrete filtering. Since the computation time is driven by the number of discrete nodes, we marginalize out previous variables each time we incorporate a new observation, maintaining a graph that contains only two nodes. We discretize the pose space by specifying a set of grasp approach directions (normal direction of the grasp surface) relative to the object, then sampling grasps on the object with 5mm of translational resolution, and 10\degree of rotational resolution. The discretized state space consists of 5k-9k poses, depending on the object size. The inference step takes 2-6 seconds per iteration, yielding a slow and coarse but global object pose signal. 
%Despite the fact that PGMax reports speedups of up to three orders of magnitude \cite{zhou2022pgmax} relative to competing solvers, inference over thousands of discrete states is very computationally expensive and takes 2-6 seconds per iteration, depending on the grid size. Therefore, we only integrate the discrete estimate every few images. This approach fortunately does not lead to much information loss due to the coarse discretization of the grid and the slow rate at which the object slides (consecutive tactile images generally correspond to the same grid point).

Fig. \ref{fig:graph}a provides insight into the architecture of the Viterbi algorithm. The variable $\mathcal{X} \in SE(2)$ represents the relative pose between the gripper and the grasped object. At the initial timestep, the environment prior is introduced. Given our prior knowledge of the environment's orientation and height, we can, for each discrete relative object pose within the grid, ascertain which point of the object would be in closest proximity to the environment and compute the corresponding distance. To do so, we transform the object pointcloud (obtained by sampling the object CAD model) by each of the poses in the grid, then save the distance of the closest point in the posed pointcloud to the ground plane in the contact normal direction. The integration of the environment prior involves the multiplication of a Gaussian function over these distances:

\begin{gather}
    \mu(\mathcal{X}_0) = P_{\text{Tac2Pose}}(\mathcal{X}_0|I_0, w_0) P_{env} (\mathcal{X}_0|g_0,c^*) \\
    P_{env} (\mathcal{X}_0|g_0,c^*) = \mathcal{N} ( p_{\text{closest}}^*(\mathcal{X}_0, g_0, c^*) \cdot \hat{n}_{c^*} ; 0, \sigma_{env})
\end{gather}

\noindent where $\mu(\mathcal{X}_0)$ is the probability of the relative gripper/object pose $\mathcal{X}_0$, $P_{\text{Tac2Pose}}(\mathcal{X}_0|I_0, w_0)$ is the single-shot estimate of probability distribution given the tactile image observation $I_0$, and the gripper width $w_0$. $P_{env} (\mathcal{X}_0|g_0)$ is the Gaussian function given the gripper pose $g_0 \in SE(2)$ and the environment prior $c^* \in SE(2)$ - the $x$-axis of $c^*$ represents the environment surface. $p_{\text{closest}}^*$ represents the closest point on the object's point cloud to the environment surface, given the relative pose $\mathcal{X}$, gripper pose $g_0$, and environment prior $c^*$, and $\hat{n}_{c^*}$ represents the unit vector normal to the environment surface. $\sigma_{env}$ determines the strength of the environment prior. In essence, the environment prior assigns higher probabilities to the relative poses that are predicted to be closer to the environment.

Subsequently, we incorporate the single-shot tactile pose estimation distribution at every $n^{th}$ step of the continuous pose estimator-controller, where $n$ is approximately five (see Fig. \ref{fig:graph}a), since the discrete pose estimator runs slower than the continuous pose estimator-controller. Instead of integrating tactile observations at a fixed frequency, we add the next tactile observation as soon as the discrete filter is ready, once the marginalization step to incorporate the previous tactile observation has been completed.

The transition probabilities impose constraints on tactile observations between consecutive time steps in the discrete graph, including:
\begin{itemize}
    \item \textit{Continuity}: The pose can transition only to neighboring poses on the pose grid to encourage continuity. (Fig. \ref{fig:viz_trasitions}-1)
    \item \textit{Persistent Contact}: The height of the closest point to the environment remains consistent across time steps due to the flat nature of the environment. This consistency is enforced through the multiplication of a Gaussian function that factors in the height difference. (Fig. \ref{fig:viz_trasitions}-2)
\end{itemize}

The first transition probability zeros out the likelihood of any transition to a non-neighboring grid point. Because the discretization of pose space is coarse, we assume the object cannot traverse more than one grid point in a single timestep. A set of allowable transitions corresponding to the first transition probability is visualized in Fig. \ref{fig:viz_trasitions}-1. 

The second transition probabilities can be mathematically expressed as follows:

\begin{align}
    P&(\mathcal{X}_i|\mathcal{X}_{i-1}) = \nonumber\\
    &\mathcal{N} ((p_{\text{closest}}^*(\mathcal{X}_i, g_i, c^*) - p_{\text{closest}}^*(\mathcal{X}_{i-1}, g_{i-1}, c^*)) \cdot \hat{n}_{c^*} ; 0, \sigma_{trs})
\end{align}

\noindent where $\sigma_{trs}$ determines the strength of this constraint. A set of transitions that are highly likely given the second transition probabilities are visualized in Fig. \ref{fig:viz_trasitions}-2.

Together, they encode the assumption that the object slides continuously within the grasp. This enables the discrete pose estimator to compute and filter the distribution of relative gripper/object poses, taking into account tactile information, robot proprioception, and environmental priors.

\begin{figure*}[h]
	\centering
	\includegraphics[width=1\linewidth]{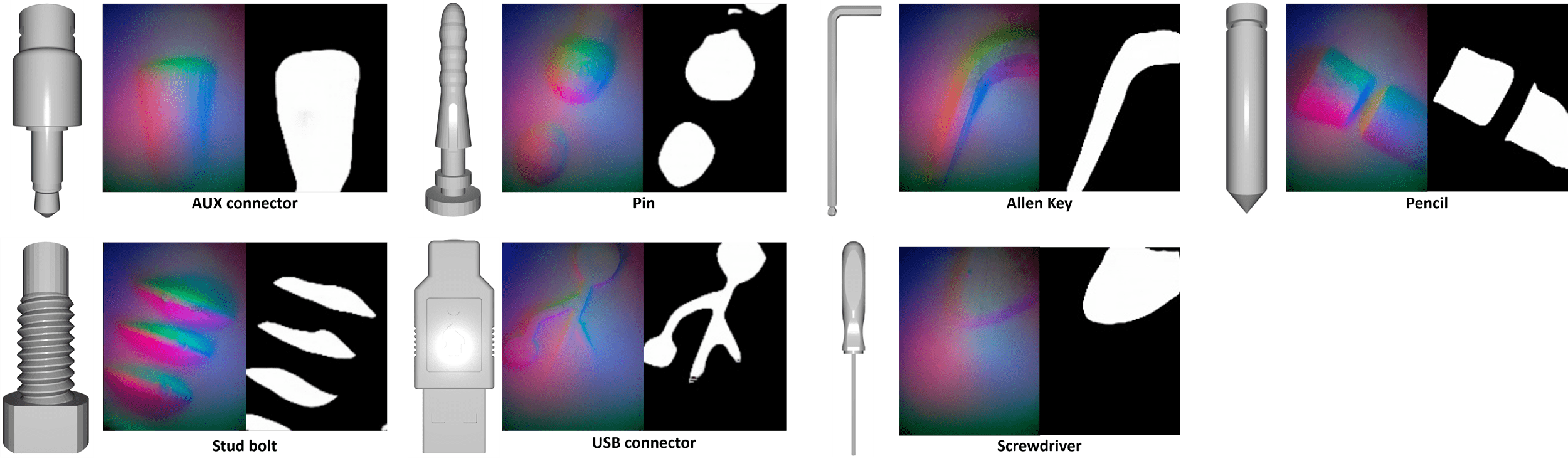}
	\caption{Test objects with example tactile images and contact patch reconstruction.}
	\label{fig:objects}
\end{figure*}

\begin{figure}[h]
	\centering
	\includegraphics[width=1\linewidth]{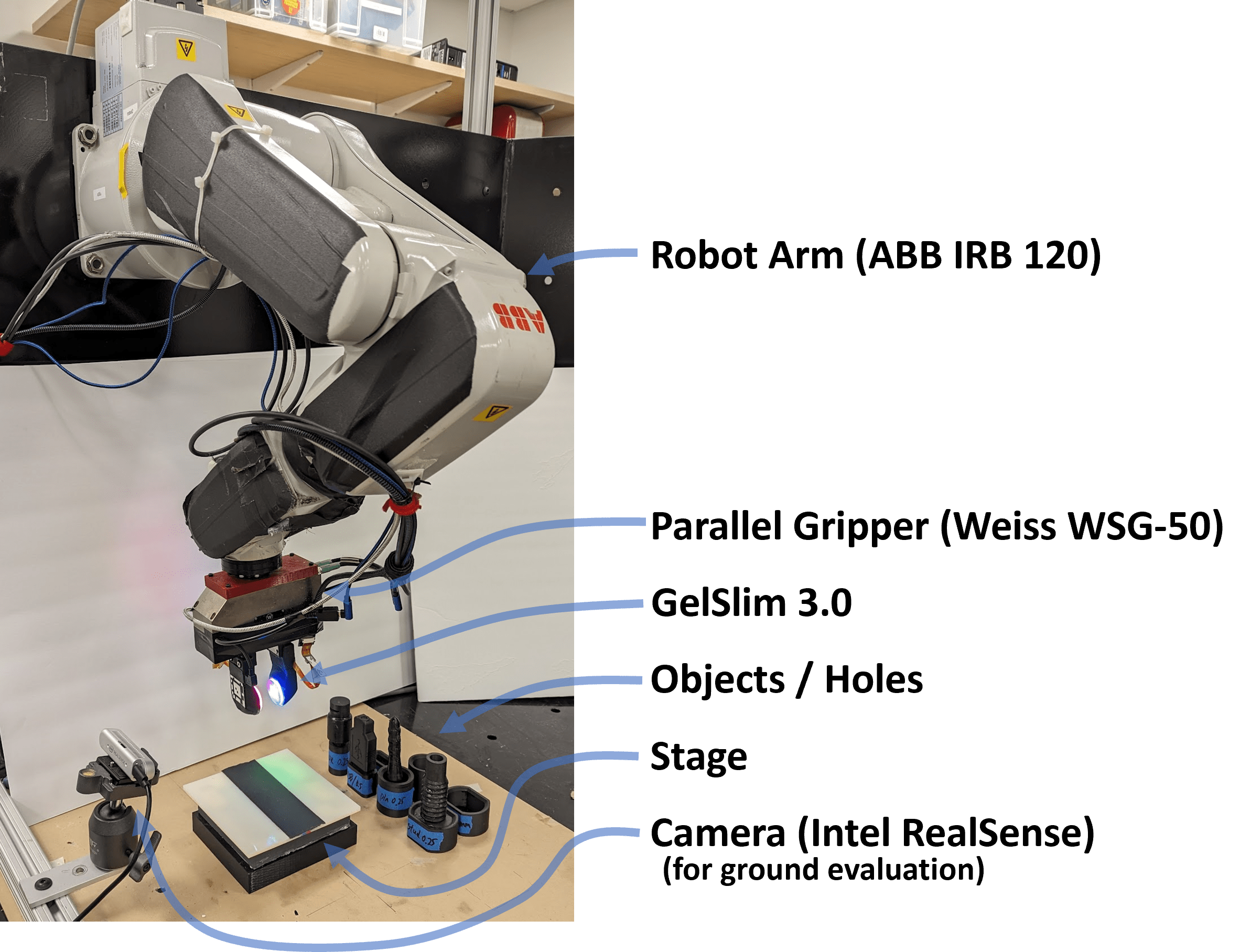}
	\caption{Hardware setup}
	\label{fig:setup}
\end{figure}

\subsection{Continuous Pose Estimator-Controller} \label{subsection:continuous}

The continuous pose estimator-controller serves a dual purpose: it takes as input the filtered discrete probability distribution of relative gripper/object poses and outputs a continuous pose estimate and an iteratively updated motion plan in a receding horizon fashion. The Incremental Smoothing and Mapping (iSAM) algorithm \cite{kaess2012isam2}, which is based on the factor graph model \cite{dellaert2012factor, dellaert2017factor}, serves as the computational backbone of our estimator-controller. We leverage its graph-based flexible formulation to combine estimation and control objectives as part of one single optimization problem.

The factor graph architecture of the continuous pose estimator-controller is illuminated in Fig. \ref{fig:graph}b. Noteworthy variables include $g_t$, $o_t$, and $c_t$, each frames in $SE(2)$, representing the gripper pose, object pose, and contact position, respectively. The orientation of $c_t$ is fixed and aligned with the normal direction of the environment. Additionally, $\mathcal{V}$ represents the set of physics parameters:
\begin{itemize}
    \item Translational-to-rotational friction ratio at the grasp: $F_{max}/M_{max}$, where $F_{max}$ and $M_{max}$ are the maximum pure force and torque that it can endure before sliding.
    \item Friction coefficient at the extrinsic contact between the object and the environment: $\mu_{max}$.
\end{itemize}

A key advantage of using the factor graph to represent the problem is that we can fuse various sources of information by formulating each piece of information as a factor. Subsequently, we can find the state that best explains the information by jointly minimizing the sum of the factor potentials, i.e. energy function. In other words, priors, measurements, kinematic constraints, physics models, and even control objectives can be represented as factors. This allows us to address both estimation and control problems simultaneously by minimizing a single energy function:

\begin{align}
    E(x) = \underbrace{\sum ||F_{\text{prior}}(\mathbf{x}_{\text{prior}})||^2}_\text{priors} + \underbrace{\sum ||F_{\text{meas}}(\mathbf{x}_{\text{meas}})||^2}_\text{measurements}  \nonumber \\
    + \underbrace{\sum ||F_{\text{cons}}(\mathbf{x}_{\text{cons}})||^2}_\text{constraints} + \underbrace{\sum ||F_{\text{model}}(\mathbf{x}_{\text{model}})||^2}_\text{models}\nonumber \\
    + \underbrace{\sum ||F_{\text{obj}}(\mathbf{x}_{\text{obj}})||^2}_\text{control objectives}
    \label{eq:squares}
\end{align}

\noindent where $F_{\text{prior}}$, $F_{\text{meas}}$, $F_{\text{cons}}$, $F_{\text{model}}$, and $F_{\text{objective}}$ are the factor potentials associated with priors, measurements, constraints, models, and control objectives, respectively. It is also noteworthy that each of the square terms is normalized by its corresponding noise model, but it is omitted for brevity. The subsets of state variables related to each factor are denoted as $\mathbf{x}_{\text{prior}}$, $\mathbf{x}_{\text{meas}}$, $\mathbf{x}_{\text{cons}}$, $\mathbf{x}_{\text{model}}$, and $\mathbf{x}_{\text{obj}}$. In Fig. \ref{fig:graph}b, each circle represents a state variable, and each dot represents a factor. The connections between variables and factors illustrate their relationships. Notably, factors labeled in red accept input from priors, measurements, or objectives, while those in grey stipulate relations between associated variables without taking any inputs.

%The framework closely resembles that of earlier work \cite{kim2023simultaneous}, where further explanation of the iSAM implementation can be found. We encourage readers to refer to this work for a description of the factors that are directly borrowed from \cite{kim2023simultaneous} ($F_{cc}, F_{oc}, F_{gp}, F_{motion}$).

The continuous estimator-controller comprises two main sections: the left segment, spanning from the initial time to the current moment $t$, is dedicated to the \textbf{estimation} of the object's pose. This estimation component considers priors, measurements, constraints, and physics models to estimate a smooth trajectory for the object's pose. The right segment, covering the time from $t$ to the control horizon $t+T$, is responsible for devising a motion plan to \textbf{control} the system and achieving the manipulation objectives. The control component takes into account constraints, physics models, and control objectives to formulate the motion plan.

In the following sections, we define each factor. The arguments of each factor definition are the variables, priors, and measurements that the factor depends on. The right-hand side specifies the quantity we are trying to optimize.

\myparagraph{Priors}

First, the environment (contact) prior is established at the initial time step:
\begin{align}
    &F_{cp}(c_1;c^*) = c^{*-1} c_1,
\end{align}
Here, $c^* \in SE(2)$ contains prior information about the environment's orientation and height. In essence, this factor penalizes the difference between the prior and the estimation. While we employ the logarithm map from the SE(2) Lie Group representation to the se(2) Lie Algebra representation to formulate the output as a three-dimensional vector, we omit the notation for brevity. This abbreviation also applies to other factors where the SE(2) transformation serves as the output.

Additionally, physics priors are imposed by formulating the factor that penalizes the difference between the prior and the estimation:
\begin{align}
    &F_{vp}(\mathcal{V};\mathcal{V^*}) = \mathcal{V} - \mathcal{V^*}.
\end{align}
where $\mathcal{V^*}$ is the prior for the physics parameters.

\myparagraph{Measurements}

The gripper pose measurement from forward kinematics ($g_i^*$) is imposed by formulating factor as difference between the measured and the estimated gripper pose:% on $g_i$:

\begin{align}
    F_{gp}(g_i;g_i^*) &:= g_i^{*-1} g_i
\end{align}

The factor graph also takes filtered pose estimations from the discrete pose estimator:
%and gripper pose measurements ($F_{gp}$).
\begin{align}
    &F_{tac}(g_i,o_i;\mathcal{X}_{i,\text{MAP}}) = \mathcal{X}_{i,\text{MAP}}^{-1} (g_i^{-1}o_i),
\end{align} 
where $\mathcal{X}_{i,\text{MAP}}$ denotes the filtered maximum a posteriori (MAP) discrete relative pose, and $(g_i^{-1}o_i)$ denotes the continuous estimate of the gripper/object relative pose. Given the higher operating speed of the continuous pose estimator-controller (0.1$\sim$0.2 seconds per iteration) compared to the discrete pose estimator (2$\sim$6 seconds per iteration), the discrete pose estimation factor is integrated when an update is available every few steps within the continuous estimator-controller. This is why we see, in Fig. \ref{fig:graph}, that this factor is not imposed at every time step.

\myparagraph{Kinematic Constraints}

Since we assume a flat environment, the location of contact on the environment should not change in the direction perpendicular to the environment surface. Additionally, the change in the tangential direction should be small, given our assumption of quasistatic motion and the absence of abrupt sliding on the environmental surface.
%if the control objective of minimizing slip is properly met.
We enforce this constraint by formulating a factor and assigning a strong noise model in the perpendicular direction and a relatively weaker noise model in the tangential direction:

\begin{align}
F_{cc}(c_{i-1},c_i) &= c_{i-1}^{-1} c_i
\end{align}

Furthermore, we assume we have 3D shape models of the objects and a prior knowledge of the normal direction of the environment. Therefore, as in the discrete filtering step, we can anticipate which part of the object would be in contact with the environment — specifically, the closest point to the environment. Consequently, we introduce a factor that incorporates the distance between the current estimated location of the contact and the closest point of the object to the environment:

\begin{align}
%F_{oc}(o_i,c_i) &= c_i^{-1} p_{\text{closest}}
F_{oc}(o_i,c_i) &= p_{\text{closest}}(o_i,c_i)
\end{align}
where $p_{\text{closest}}(o_i,c_i)$ represents the point in the object's point cloud that is closest to the environment direction, expressed in the contact frame $c_i$.

\myparagraph{Physics Model}

We impose a friction model based on the limit-surface model \cite{goyal1989planar,chavan2018hand} as a transition model to capture the dynamics of sliding ($F_{fric}$). This model provides a relation between the kinetic friction wrench and the direction of sliding at the grasp. In essence, it serves as a guide for predicting how the object will slide in response to a given gripper motion and extrinsic contact location. The relation is formally represented as follows:
\begin{align}
    [\omega, v_x, v_y] \propto [\frac{M}{M_{max}^2}, \frac{F_x}{F_{max}^2}, \frac{F_y}{F_{max}^2}]. \label{eq:limitsurface}
\end{align}
Here, $[\omega, v_x, v_y]$ denotes the relative object twist in the gripper's frame, i.e. sliding direction, while $[M, F_x, F_y]$ signifies the friction wrench at the grasp. To fully capture the friction dynamics, additional kinematic and mechanical constraints at the extrinsic contact are also considered. These constraints are formulated as follows:
\begin{gather}
    M\hat{z} - \vec{l}_{gc} \times \vec{F} = 0, \label{eq:torquebal}\\%  \text{  (zero torque at the extrinsic contact)},\\
    v_{c, N}(g_{i-1},o_{i-1},c_{i-1},g_i,o_i) = 0, \label{eq:constnormal}\\
    v_{c, T}(g_{i-1},o_{i-1},c_{i-1},g_i,o_i) = 0 \qquad \label{eq:consttangent}\\ \qquad \perp (F_T = -\mu_{max} F_N \ \text{OR} \ F_T = \mu_{max} F_N ), \label{eq:frictioncone}
\end{gather}
In these equations, $\vec{l}_{gc}$  is the vector from the gripper to the contact point, and $v_{c, N}$ and $v_{c, T}$ represent the local velocities of the object at the point of contact in the directions that are normal and tangential to the environment, respectively. $F_N$ and $F_T$ denote the normal and tangential components of the force. Eq. \ref{eq:torquebal}  specifies that no net torque should be present at the point of extrinsic contact since we are assuming point contact. Eq. \ref{eq:constnormal} dictates that the normal component of the local velocity at the point of extrinsic contact must be zero as long as contact is maintained. Eq. \ref{eq:consttangent} and Eq. \ref{eq:frictioncone} work complementarily to stipulate that the tangential component of the local velocity at the contact point must be zero (Eq. \ref{eq:consttangent}), except in cases where the contact is sliding. In such instances, the contact force must lie on the boundary of the friction cone (Eq. \ref{eq:frictioncone}). By combining Eq. \ref{eq:limitsurface}$\sim$\ref{eq:frictioncone}, we establish a fully determined forward model for the contact and object poses, which allows the object pose at step $i$ to be expressed as a function of its previous poses, the current gripper pose, and the physics parameters:
\begin{equation}
    o_i^* = f(g_{i-1},o_{i-1},c_{i-1},g_i,\mathcal{V})
\end{equation}
This relationship can thus be encapsulated as a friction factor:
\begin{align}
    &F_{fric}(g_{i-1},o_{i-1},c_{i-1},g_i,o_i,\mathcal{V}) = o_i^{*-1}o_i.
\end{align}

With all the previously introduced factors combined, the estimation component formulates a smooth object pose trajectory that takes into account priors, tactile measurements, robot kinematics, and physics model.

\myparagraph{Control Objective}

The control segment incorporates multiple auxiliary factors to facilitate the specification of regrasping objectives. First, the desired goal configuration is imposed at the end of the control horizon ($F_{goal}$). This comprises three distinct sub-factors, corresponding to the three manipulation objectives described in Section \ref{formulation}, which can be turned on or off, depending on the desired configuration:
\begin{enumerate}
    \item \( F_{\text{goal,go}} \) regulates the desired gripper/object relative pose at \( o_{t+T} \) and \( g_{t+T} \).
    \item \( F_{\text{goal,o}} \) enforces the object's orientation within the world frame at \( o_{t+T} \).
    \item \( F_{\text{goal,c}} \) dictates the desired contact point at \( c_{t+T} \), thereby facilitating controlled sliding interactions with the environment.
\end{enumerate}
These sub-factors are mathematically expressed as follows:
\begin{align}
    F_{goal,go}(g_{t+T},o_{t+T}) &= p_{o,goal}^{g \ -1} (g_{t+T}^{-1} o_{t+T}),\\
    F_{goal,o}(o_{t+T}) &= o_{goal}^{-1} o_{t+T},\\
    F_{goal,c}(c_{t+T}) &= c_{goal}^{-1} c_{t+T}.
\end{align}
Here, $p_{o,goal}^{g}$ signifies the target relative gripper/object pose, \( o_{goal} \) represents the desired object orientation in the world frame, and \( c_{goal} \) is the intended contact point.

Additionally, the \( F_{motion} \) factor minimizes the gripper motion across consecutive time steps to reduce redundant motion and optimize for a smooth gripper trajectory.

\begin{align}
    F_{motion}(g_{i-1},g_i) = g_{i-1}^{-1} g_i
\end{align}

Concurrently, a contact maintenance factor, \( F_{cm} \), serves as a soft constraint to direct the gripper's motion in a way that prevents it from losing contact with the environment:
\begin{align}
    F_{cm}(g_{i-1}, &c_{i-1}, g_i; \epsilon_i) = max(0, \zeta_i(g_{i-1}, c_{i-1}, g_i) + \epsilon_i),
    %&= max\left(0, \left(c_{i-1}^{-1} (g_{i-1}^{-1} g_{i} g_{i-1}^{-1} c_{i-1})\right)_y - \epsilon_i \right).
\end{align}
where \( \zeta_i \) represents the normal component of the virtual local displacement from step  \( i-1 \) to \( i \) at the contact point, assuming the grasp is fixed. The term \( \epsilon_i \) is a small positive scalar, encouraging \( \zeta_i \) to be negative, thus fostering a motion that pushes against the environment.

Taken together, these factors cohesively formulate a motion plan from $g_{t+1}$ to $g_{t+T}$, which is then communicated to the robot. The robot continues to follow the interpolated trajectory of this motion plan until it receives the next update, akin to model predictive control.

\begin{figure*}[h]
	\centering
	\includegraphics[width=1.0\linewidth]{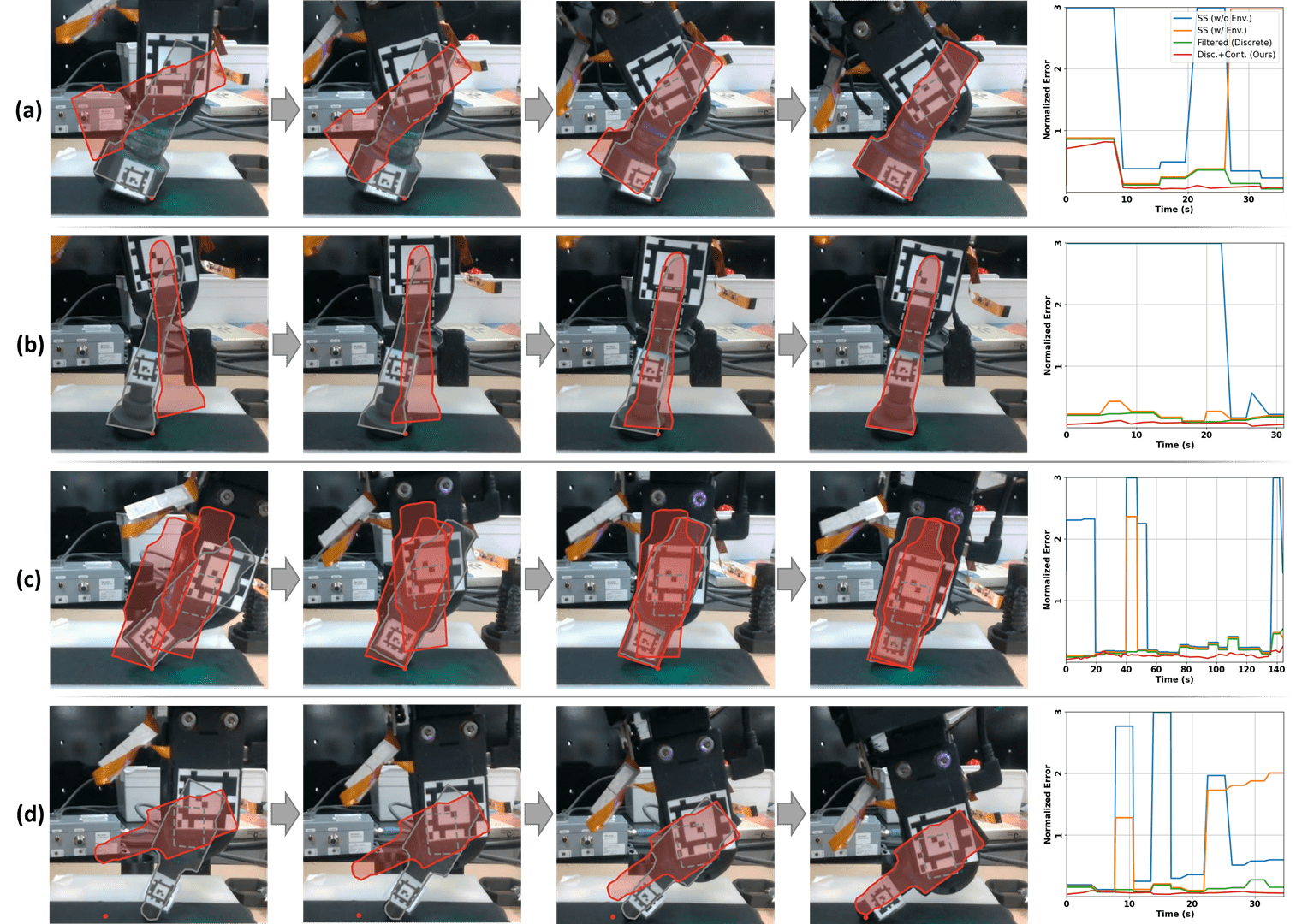}
	\caption{Demonstrations of four types of goal configurations: (a) Relative Orientation + Stationary Extrinsic Contact, (b) Relative Orientation/Translation + Stationary Extrinsic Contact, (c) Relative Orientation + Global Orientation + Stationary Extrinsic Contact, and (d) Relative Orientation + Sliding Extrinsic Contact. The right column depicts normalized estimation accuracy for the proposed method and ablation models.}
	\label{fig:samples}
\end{figure*}

\subsection{Ablation Models}

We implemented five ablation models to investigate the contribution of specific components to estimation accuracy:

\begin{itemize}
    \item Environment priors (height/orientation)
    \item Multi-shot filtering (v.s. single-shot estimate)
    \item Continuous estimation (v.s. discrete estimate)
    \item Quality of contact patch reconstruction
    \item Accuracy of the physics prior
\end{itemize}

\myparagraph{SS (w/o Env.):} This model, equivalent to the previous Tac2Pose algorithm \cite{bauza2022tac2pose}, uses a single tactile image and the gripper width to compute the probability distribution of relative poses between the gripper and the grasped object ('Single-shot Tactile Pose Estimate' in Fig. \ref{fig:overview}).

\myparagraph{SS (w/ Env.):} In addition to the tactile image and gripper width, this model incorporates priors on the height and orientation of the environment floor. Comparing this model with 'SS (w/o Env.)' provides insights into the contribution of the environment priors to estimation accuracy. All subsequent ablation models incorporate environment priors.

\myparagraph{Filtered (Discrete):} This model utilizes the discrete filter to fuse a stream of multiple tactile images ('Filtered Tactile Pose' in Fig. \ref{fig:overview}). Comparing with 'SS (w/ Env.)' helps assess the impact of fusing multiple tactile images on accuracy compared to using just a single tactile image.

\myparagraph{Discrete+Continuous (Ours):} Our proposed model. The following two ablation models leverage privileged information to evaluate potential improvements in estimation accuracy.

\myparagraph{Discrete+Continuous (Privileged):} This model uses privileged information to synthesize the binary contact patch. From the Apriltag attached to the grasped object, it computes the ground truth relative pose between the gripper and the object. Based on the relative pose, it synthesizes the anticipated binary contact patch rather than inferring it from actual tactile images. Since the same contact patch synthesis method was used during the training of the Tac2Pose model, this model shows how the system would perform if the binary contact patch reconstruction were the same as the ground truth.

\myparagraph{Discrete+Continuous (Simulation):} This model provides insights into the system's performance under the assumption of an exact physics prior. The methodology involves simulating the object trajectory based on the identical gripper trajectory used in other models. The object trajectory simulation employs a modified factor graph. By imposing only the priors, kinematic constraints, gripper motion, and the physics model, we can find the object trajectory that exactly aligns with the physics model. Consequently, using the same physics parameters for this simulation factor graph as those in our prior and synthesizing the contact patch corresponding to the simulated object trajectory allows us to evaluate how effectively our system would perform with both the exact physics model and contact patch reconstruction.

\section{Experiments and Results}

\begin{table*}[h]
\caption{Median Normalized Estimation Errors}
\centering
\begin{tblr}{hlines}
& {AUX\\(6 trajectories)} & {Pin\\(5 trajectories)} & {Stud\\(3 trajectories)} & {USB\\(4 trajectories)} & {\textbf{Overall}\\(18 trajectories)} \\
\hline
{SS (w/o Env.)}      & 0.92 & 2.00 & 1.47 & 1.12 & 1.41 \\
{SS (w/ Env.)}      & 0.21 & 0.12 & 0.30 & 0.25 & 0.20 \\
{Filtered (Discrete)} & 0.17 & 0.10 & 0.18 & 0.17 & 0.15 \\
{\textbf{Discrete+Continuous (Ours)}}    & \textbf{0.10} & \textbf{0.07} & \textbf{0.07} & \textbf{0.07} & \textbf{0.07} \\
{Discrete+Continuous (Privileged)} & 0.06 & 0.09 & 0.08 & 0.07 & 0.07 \\
{Discrete+Continuous (Simulation)}  & 0.03 & 0.05 & 0.10 & 0.06 & 0.05 \\
\end{tblr}
\label{table:estimation}
\end{table*}

We conducted a series of experiments on four distinct 3D-printed objects and three household items (illustrated in Fig. \ref{fig:objects}) to validate the efficacy of our algorithm. The experiments were designed to:

\begin{enumerate}
\item Quantitatively evaluate the algorithm's performance across a variety of target configurations.
\item Qualitatively demonstrate the utility of the algorithm with household items in various scenarios.
\item Assess the algorithm's applicability to specific real-world tasks, such as object insertion.
\end{enumerate}

\subsection{Experimental Setup} 

Fig. \ref{fig:setup} shows the hardware setup, which includes a 6-DoF ABB IRB 120 robot arm, Weiss WSG-50 parallel gripper, and a GelSlim 3.0 sensor \cite{taylor2022gelslim}. On the table, there is a stage that serves as a flat environment, as well as objects and holes for the insertion experiment. Additionally, there is an Intel RealSense camera used to track the object's pose through Apriltags attached to the objects to obtain the ground truth pose. The Apriltag attached to the gripper serves the purpose of calibration.

\begin{figure}[t]
	\centering
	\includegraphics[width=1\linewidth]{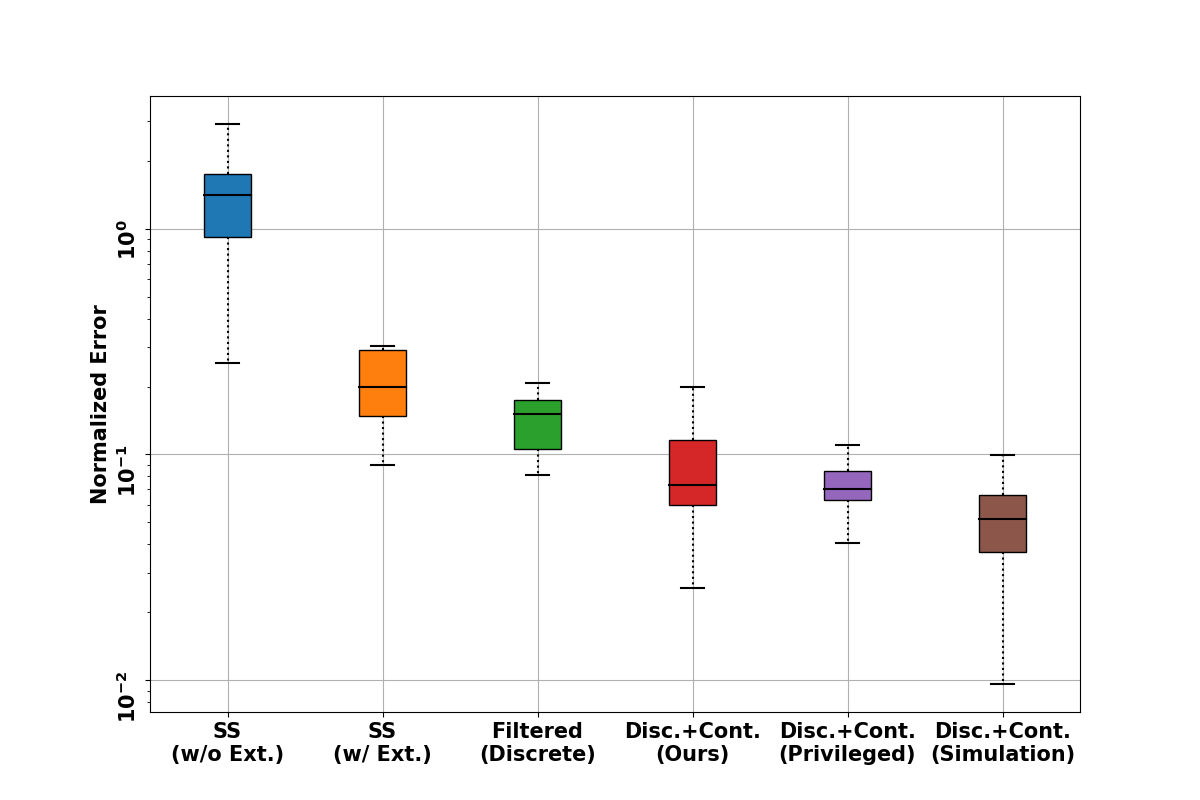}
	\caption{Normalized Estimation Errors.}
	\label{fig:err}
\end{figure}

\subsection{Performance Across Various Goal Configurations} \label{subsection:varigoal}

We assessed our algorithm's performance using a total of 18 diverse goal configurations. Our framework allows for specifying goals relative to the gripper (regrasping) and relative to the world frame (reorienting), facilitating different downstream tasks. For example, regrasping can improve grasp stability, enable tactile exploration, and establish a grasp optimized for both force execution and the avoidance of collisions or kinematic singularities in downstream tasks. On the other hand, reorienting the object can enable mating with target objects in the environment or prevent collisions with obstacles. The configurations we evaluate fall into four distinct categories:

\begin{itemize}
    \item Relative Orientation + Stationary Extrinsic Contact
    \item Relative Orientation/Translation + Stationary Extrinsic Contact
    \item Relative Orientation + Global Orientation + Stationary Extrinsic Contact
    \item Relative Orientation + Sliding Extrinsic Contact
\end{itemize}

Examples of these four goal configuration types are illustrated in Fig. \ref{fig:samples}, along with corresponding plots showcasing estimation accuracy. The red silhouettes that move along with the gripper represent the desired relative pose between the gripper and the object. Conversely, the grey silhouettes depict object poses as measured by Apriltags, which we use as the ground truth object pose. The red dots mark the desired extrinsic contact location. In Fig. \ref{fig:samples}c, the other red silhouette signifies the desired object orientation in the global frame. The time series plots on the right column indicate the performance of the proposed and ablation models. These results attest to the algorithm's adeptness in attaining desired goal configurations while showing better estimation performance compared to ablation models.

A summary of each algorithm's estimation performance is presented in Fig. \ref{fig:err}; the error per-object is broken out in Table \ref{table:estimation}. The error values denote the normalized estimation error, computed as follows:

\begin{equation}
    \epsilon_{\text{norm}} = ||(\epsilon_{\text{rot}}, \epsilon_{\text{trn}} / (l_\text{obj}/2)||_1
\end{equation}

Here, $||\cdot||_1$ signifies the L1-norm, $\epsilon_{\text{rot}}$ indicates rotation error in radians, $\epsilon_{\text{trn}}$ denotes translation error, and $l_\text{obj}$ represents the object's length. In essence, this value signifies the overall amount of estimation error normalized by objects' size. This analysis reveals how much each system component contributes to the estimation accuracy.

\subsubsection{Effect of Environment Priors}

Firstly, there is a substantial decrease in normalized error from 1.41 to 0.20 when transitioning from 'SS (w/o Env.)' to 'SS (w/ Env.)'. Without environment priors – no information about the height and orientation of the environment – the estimator suffers due to ambiguity in tactile images, as thoroughly explored in \cite{bauza2022tac2pose}. For most grasps of the objects we experiment with, a single tactile imprint is not sufficient to uniquely localize the object. For instance, the local shape of the pin and the stud exhibits symmetry, making it challenging to distinguish if the object is held upside-down, resulting in a very high estimation error with the 'SS (w/o Env.)' model. In contrast, the 'SS (w/ Env.)' model was able to significantly resolve this ambiguity by incorporating information about the environment.

This suggests that knowing when object is in contact with a known environment can be used effectively to collapse the ambiguity in a single tactile imprint. Although this knowledge, on its own, is a weak signal of pose, it provides global context that, when paired with a tactile imprint, can yield accurate pose estimation. Much prior work prefers vision as a modality to provide global pose context; this analysis demonstrates that prior knowledge of the object and environment (when available) can be leveraged to provide global context instead of introducing additional sensors and algorithms.

\subsubsection{Effect of Multi-shot Filtering}

Between the 'SS (w/ Env.)' and the 'Filtered (Discrete)' models, the normalized error significantly decreases from 0.20 to 0.15. This shows that fusing a stream of multiple tactile images is effective in improving the estimation accuracy. The discrete filter is able to reduce ambiguity by fusing information over a sequence of tactile images, obtained by traversing the object surface and therefore exposing the estimator to a more complete view of the object geometry. Fusing information over multiple tactile images also robustifies the estimate against noise in the reconstruction of any individual contact mask. The difference is distinctive in the time plots of the normalized error in Fig. \ref{fig:samples}. While the 'SS (w/ Env.)' and the 'Filtered (Discrete)' model have an overlapping error profile for the majority of the time, there is a significant amount of portion where the errors of the 'SS (w/ Env.)' model suddenly surges. This is because the 'SS (w/ Env.)' model only depends on a single tactile image snapshot, and therefore does not consider the smoothness of the object pose trajectory over time. In contrast, the error profile of the 'Filtered (Discrete)' model is smoother since it considers consistency in the object pose.

\subsubsection{Effect of Continuous Estimation}

The median normalized error also decreases from 0.15 of the 'Filtered (Discrete)' model to 0.07 of the 'Discrete+Continuous (Ours)' model. This improvement is attributed to the continuous factor graph refining the discrete filtered estimation with more information in both spatial and temporal resolution. While the discrete filter runs at a lower frequency, the continuous factor graph operates at a higher frequency. This means that it takes in gripper pose measurements even when the discrete estimate from the tactile image is not ready. Additionally, it considers the physics model when computing the estimate. Consequently, the 'Discrete+Continuous (Ours)' model results in a smoother and more physically realistic trajectory estimate, as evident in the error time plots in Fig. \ref{fig:samples}.

\subsubsection{Potential Effect of Ground Truth Contact Patch Reconstruction}

Fig. \ref{fig:err} suggests that the difference between 'Discrete+Continuous (Ours)' and 'Discrete+Continuous (Privileged)' is not significant. This implies that having the ground truth contact patch reconstruction would not significantly improve the accuracy of the estimation. It suggests that the contact patch reconstruction has sufficiently good quality, retaining significant information compared to the ground truth contact patch. This is attributed to the significant domain randomization incorporated into the contact patch reconstruction during Tac2Pose model training.

When training the Tac2Pose model, the input is the binary contact patch, and the output is the probability distribution of contact poses. The training data for the binary contact patch are synthesized using the local 3D shape of the object model. To overcome the sim-to-real gap in contact patch reconstruction, random errors are intentionally introduced to the synthesized contact patch as discussed in Section \ref{subsection:discrete}. This result indicates that, thanks to effective domain randomization, the model does not suffer significantly from the sim-to-real gap in contact patch reconstruction.

\subsubsection{Potential Effect of the Exact Physics Model}

Fig. \ref{fig:err} shows a significant decrease in normalized error when using the simulated physics that exactly aligns with our physics prior. A noteworthy observation is that it does not reduce the normalized error to zero, indicating that our estimation would still not be perfect even with the exact physics model. This aligns with intuition, as tactile observation is a local observation and cannot guarantee full observability even when we know the physics exactly.

\begin{figure*}[h]
	\centering
	\includegraphics[width=1.0\linewidth]{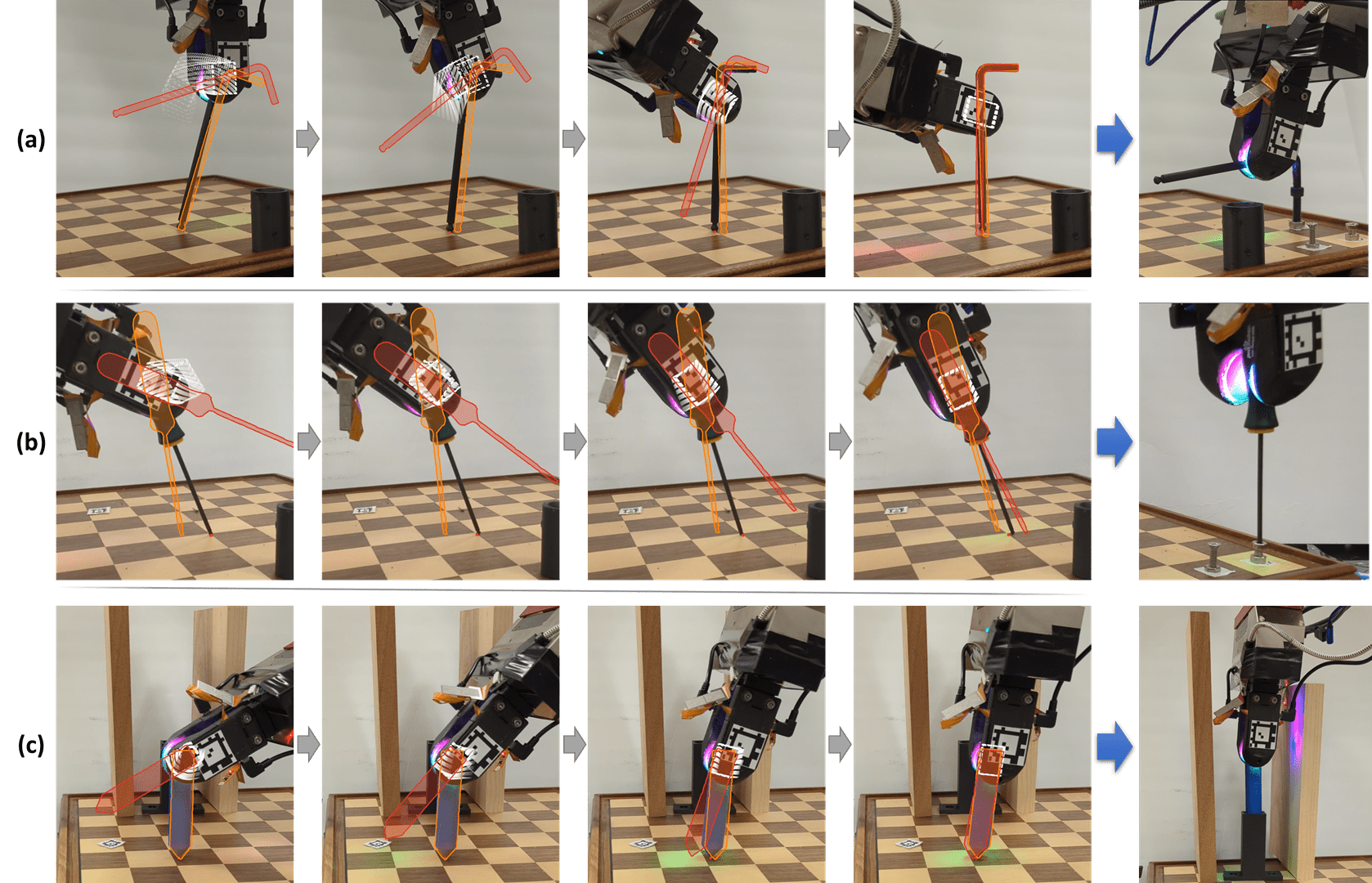}
	\caption{Demonstrations with household items in various scenarios: (a) Adjusting the grasp of the Allen key to exert a sufficient amount of torque when screwing a bolt, (b) Adjusting the grasp of the screwdriver to prevent hitting the robot’s motion limit while screwing a bolt, (c) Adjusting the grasp of the pencil to ensure the robot does not collide with obstacles when placing the pencil in the pencil holder.}
	\label{fig:qual}
\end{figure*}

\subsection{Demonstration with Household Items in Various Scenarios} \label{subsection:household}

We additionally demonstrate our algorithm with real objects in realistic scenarios that we would face in daily life (Fig. \ref{fig:qual}):

\begin{itemize}
    \item \textit{Allen Key}: Adjusting the grasp of the Allen key to exert a sufficient amount of torque when screwing a bolt.
    \item \textit{Screwdriver}: Adjusting the grasp of the screwdriver to prevent hitting the robot's motion limit or singularity while screwing a bolt.
    \item \textit{Pencil}: Adjusting the grasp of the pencil to ensure the robot does not collide with obstacles when placing the pencil in the pencil holder.
\end{itemize}

The left four columns of Fig. \ref{fig:qual} are the snapshots of the motions over time. The rightmost column of the figure illustrates the downstream tasks after the regrasp is done. In the figure, the red silhouettes illustrate the relative goal grasp, manually selected based on the downstream task we want to achieve. The orange silhouettes represent the current estimate of the object pose. The white superimposed rectangles illustrate the planned motion trajectories of the gripper to achieve the desired configurations.

\subsubsection{Allen Key (Fig. \ref{fig:qual}a)}

In the Allen Key example, the initial grasp is on the corner part of the object, making it challenging to exert a sufficient amount of torque. Therefore, we adjusted the grasp by imposing the goal grasp pose on the longer side of the Allen Key. This allows for a longer torque arm length, ensuring the robot can exert a sufficient amount of torque. The orientation of the goal grasp was also set to keep the robot's motion within the feasible range during the screwing process.

A notable observation is that the algorithm first attempts to pivot the object before pushing it down against the floor to slide the grasp. This suggests that the algorithm effectively considers the physics model both on the finger and the floor to plan for a reasonable and intuitive motion. Without incorporating such a physics model, the motion could result in counterintuitive movements, potentially causing the object to slip on the floor.

\subsubsection{Screwdriver (Fig. \ref{fig:qual}b)}

In the Screwdriver example, the initial grasp is configured such that the screwing axis and the robot wrist axis are not aligned. This configuration could lead to issues when attempting to screw at a large angle, as it requires a more extensive motion of the robot arm compared to when the screwing axis and the wrist axis are aligned. Additionally, it may cause the robot arm to reach its motion limit. Conversely, by regrasping the screwdriver and aligning the screwing axis with the wrist axis, the robot can easily screw the bolt with primarily wrist axis rotation. Therefore, we set the goal grasp pose to align the screwing axis and the wrist axis.

While the algorithm was able to get close to the goal grasp, the estimation error was significantly larger than in the other two cases. This is because the screwdriver has less distinctive tactile features than the other items. In Fig. \ref{fig:objects}, we can see that the screwdriver has an oval-shaped contact patch without straight lines or sharp corners. In contrast, the Allen key and the pencil exhibit more distinctive straight lines and sharp corners. Since these distinctive features are crucial for resolving estimation uncertainty, the screwdriver shows less estimation accuracy than the other two.

\subsubsection{Pencil (Fig. \ref{fig:qual}c)}

In the pencil example, the robot wrist axis and the pencil are not aligned in the initial grasp. Given the obstacles next to the pencil holder, the robot would likely collide with the obstacles without adjusting the grasp. Therefore, we set the goal grasp to enable the robot to avoid collisions with the obstacles. Consequently, the robot achieved an appropriate grasp while keeping track of the object pose estimate, and then successfully placing the pencil in the holder without colliding with obstacles.

\subsection{Practical Application: Insertion Task}

\begin{table}[h]
\caption{Insertion Experiment Results (Success/Attempt)}
\centering
\begin{tblr}{hlines}
Clearance & AUX     & Pin    & Stud   & USB     \\ 
\hline
1 mm   & 10 / 10 & 6 / 10 & 7 / 10 & 7 / 10  \\ 
0.5 mm & 9 / 10  & 3 / 10 & 5 / 10 & 6 / 10  \\
\end{tblr}
\label{table:insertion}
\end{table}

To validate our algorithm's practical utility, we applied it to a specific downstream task — object insertion with small clearance (1$\sim$0.5 mm). For these experiments, we sampled random goal configurations from the first category (adjusting relative orientation) described in Section \ref{subsection:varigoal}. Following this, we aimed to insert the grasped object into holes with 1 mm and 0.5 mm total clearance in diameter.

Table \ref{table:insertion} summarizes the outcomes of these insertion attempts. The AUX connector, which features a tapered profile at the tip, had a success rate exceeding 90\%. On the other hand, the success rate dropped considerably for objects with untapered profiles, especially when the clearance was narrowed from 1 mm to 0.5 mm. The varying performance is consistent with our expectations, given that the algorithm's median normalized pose estimation error is 0.07, which corresponds to approximately 2$\sim$3 mm of translation error as quantified in Section \ref{subsection:varigoal}.

These findings indicate that our algorithm is useful in tasks that necessitate regrasping and reorienting objects to fulfill downstream objectives by meeting the goal configuration. However, for applications requiring sub-millimeter accuracy, the algorithm's performance would benefit from integration with a compliant controlled insertion policy (e.g., \cite{inoue2017deep, dong2021tactile, zhao2022offline, kim2022active}).

\section{Conclusion}

This paper introduces a novel simultaneous tactile estimator-controller tailored for in-hand object manipulation. The framework harnesses extrinsic dexterity to regrasp a grasped object while simultaneously estimating object poses. This innovation holds particular promise in scenarios necessitating object or grasp reorientation for tasks like insertion or tool use, particularly in cases where the precise visual perception of the object's global pose is difficult due to occlusions. 

We show the capability of our algorithm to autonomously generate motion plans for diverse goal configurations that encompass a range of manipulation objectives, then execute them precisely via high-accuracy tactile pose estimation (approximately 2$\sim$3mm of error in median) and closed-loop control. We further demonstrate the practical utility of our approach in solving high-tolerance insertion tasks, as well as showcase our method's ability to generalize to household objects in realistic scenarios, encompassing a variety of material, inertial, and frictional properties.

In future research, our focus will extend to investigating methodologies for autonomously determining optimal target configurations for task execution, eliminating the need for manual specification. Additionally, we are keen on exploring the potential of inferring physics parameters online or integrating a more advanced physics model capable of reasoning about the intricacies of real-world physics.

%\addtolength{\textheight}{-12cm}

\bibliographystyle{IEEEtran}
\bibliography{ref}

\end{document}